\documentclass[letterpaper, 10 pt, conference]{ieeeconf}  % Comment this line out if you need a4paper

\IEEEoverridecommandlockouts                              % This command is only needed if 
\usepackage{graphics} % for pdf, bitmapped graphics files
\usepackage{epsfig} % for postscript graphics files
\usepackage{mathptmx} % assumes new font selection scheme installed
\usepackage{times} % assumes new font selection scheme installed
\usepackage{amsmath} % assumes amsmath package installed
\usepackage{amssymb}  % assumes amsmath package installed
\usepackage{algorithm}
\usepackage{algorithmic}
\usepackage{gensymb}
\usepackage{multirow}
\usepackage{subcaption}
\usepackage{cite}
\usepackage{tikz}
\usetikzlibrary{shapes,arrows,chains}
\usepackage{pgfplots}

\makeatletter
\let\NAT@parse\undefined
\makeatother
\usepackage{hyperref}

\usepackage{color}

\title{\LARGE \bf
VRSO: Visual-Centric Reconstruction for Static Object Annotation
}

\author{\normalsize Chenyao Yu$^{1*}$, Yingfeng Cai$^{1*}$, Jiaxin Zhang$^{1*}$, Hui Kong$^{2}$, Wei Sui$^{1}$ and Cong Yang$^{1\dag}$}%

\begin{document}
\twocolumn[{
\renewcommand\twocolumn[1][]{#1}
\maketitle

\if 0
\begin{center}   
    \centering
    \includegraphics[width=1.0\linewidth]{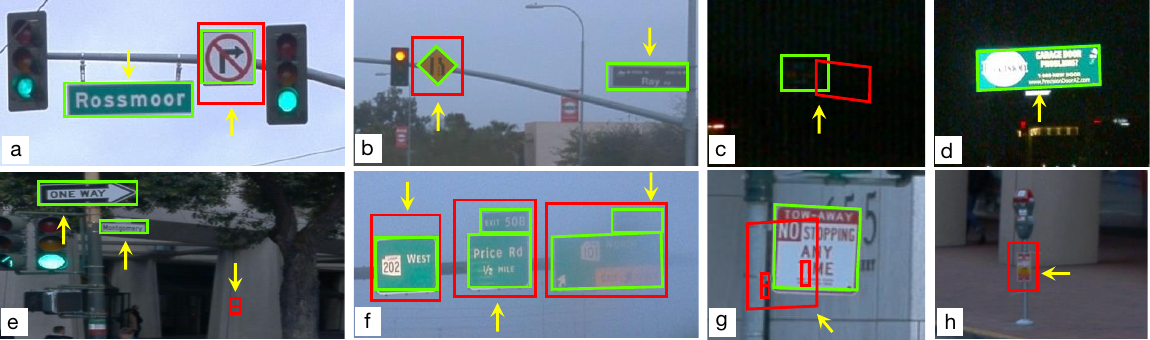}
    \captionof{figure}{\small The reprojection of 3D annotation in image space. Reprojected images are zoomed in to show a detailed comparison of annotation from our method (in green boxes) and Waymo Open Dataset (WOD) manual labeling (in red boxes). It can be seen that the annotation of WOD is not always consistent with the 2D images. It also contains some false positives (FP) and false negatives (FN) results. For example, there are signs in the image but annotation from WOD did not show positive results, (a, c). The reprojection annotation in (b) shows that the signboard is not visible in the image. In contrast, VRSO can provide more accurate and consistent 3D annotation in different cases even in poor image quality and difficult illuminance conditions(d, e, f).}
   \label{fig:teaser}
\end{center}
\fi
\begin{center}   
    \centering
    \includegraphics[width=1.0\linewidth]{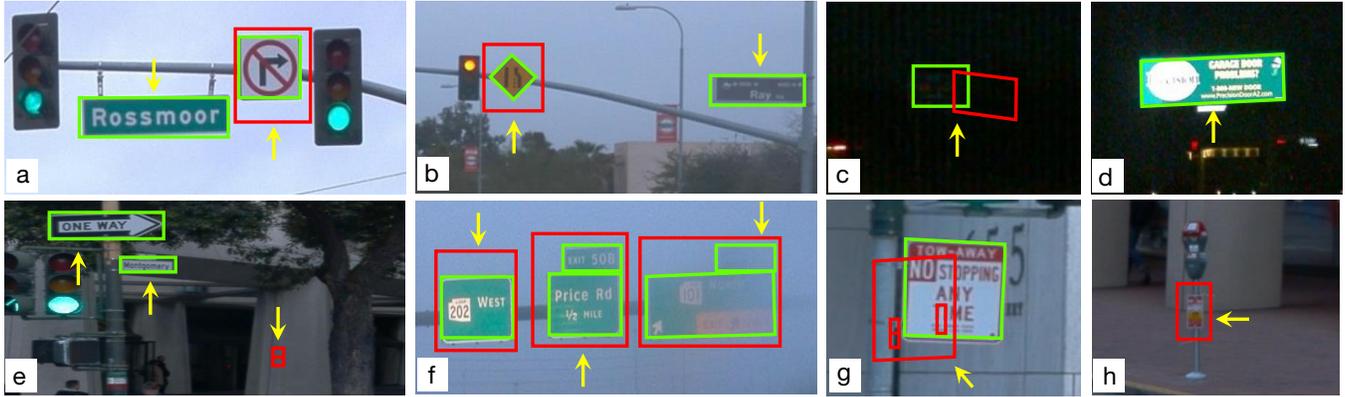}
    \captionof{figure}{\small Comparison between our proposed VRSO (green) and Waymo (red) annotations after reprojection from 3D space to 2D images. All images are from the Waymo Open Dataset (WOD). We can easily find the reprojection errors (false positives and false negatives, marked by yellow arrows) among the WOD annotations. For instance, the traffic signs are (1) missed in (a), (b), (d) and (e), (2) incorrectly labelled in (c), (e), and (h),  (3) not tightly covered in (a), (b), (f) and (g). Differently, VRSO yields consistent and accurate annotation among all images, even in low-resolution (e), illuminate conditions (c) and (d). Best viewed in colour.}
   \label{fig:teaser}
\end{center}

}]

\let\thefootnote\relax\footnotetext{
$\dag$ Corresponding author: Cong Yang (cong.yang@suda.edu.cn).
Affiliation:
$^{1}$ Ecology and Innovation Center of Intelligent Driving (BeeLab), School of Future Science and Engineering, Soochow University, Suzhou, China.
$^{2}$ Faculty of Science and Technology, University of Macau, Taipa, Macao SAR. 
This work was funded by the National Natural Science Foundation of China (Grant Number: 62473276), the Natural Science Foundation of Jiangsu Province (Grant Number: BK20241918), the Natural Science Foundation of the Jiangsu Higher Education Institutions of China (Grant Number: 22KJB520008), the Research Fund of Horizon Robotics (H230666), and the Jiangsu Policy Guidance Program, International Science and Technology Cooperation, The Belt and Road Initiative Innovative Cooperation Projects (BZ2021016). *: Equal contribution.
}

%%%%%%%%%%%%%%%%%%%%%%%%%%%%%%%%%%%%%%%%%%%%%%%%%%%%%%%%%%%%%%%%%%%%%%%%%%%%%%%%
\begin{abstract}
As a part of the perception results of intelligent driving systems, static object detection (SOD) in 3D space provides crucial cues for driving environment understanding. With the rapid deployment of deep neural networks for SOD tasks, the demand for high-quality training samples soars. The traditional, also reliable, way is manual labelling over the dense LiDAR point clouds and reference images. Though most public driving datasets adopt this strategy to provide SOD ground truth (GT), it is still expensive and time-consuming in practice. This paper introduces VRSO, a visual-centric approach for static object annotation. Experiments on the Waymo Open Dataset show that the mean reprojection error from VRSO annotation is only 2.6 pixels, around four times lower than the Waymo Open Dataset labels (10.6 pixels). VRSO is distinguished in low cost, high efficiency, and high quality: (1) It recovers static objects in 3D space with only camera images as input, and (2) manual annotation is barely involved since GT for SOD tasks is generated based on an automatic reconstruction and annotation pipeline.
\end{abstract}

%%%%%%%%%%%%%%%%%%%%%%%%%%%%%%%%%%%%%%%%%%%%%%%%%%%%%%%%%%%%%%%%%%%%%%%%%%%%%%%%
\section{INTRODUCTION}
\label{sec:intro}
\begin{figure}[t!]
    \includegraphics[width=0.49\textwidth]{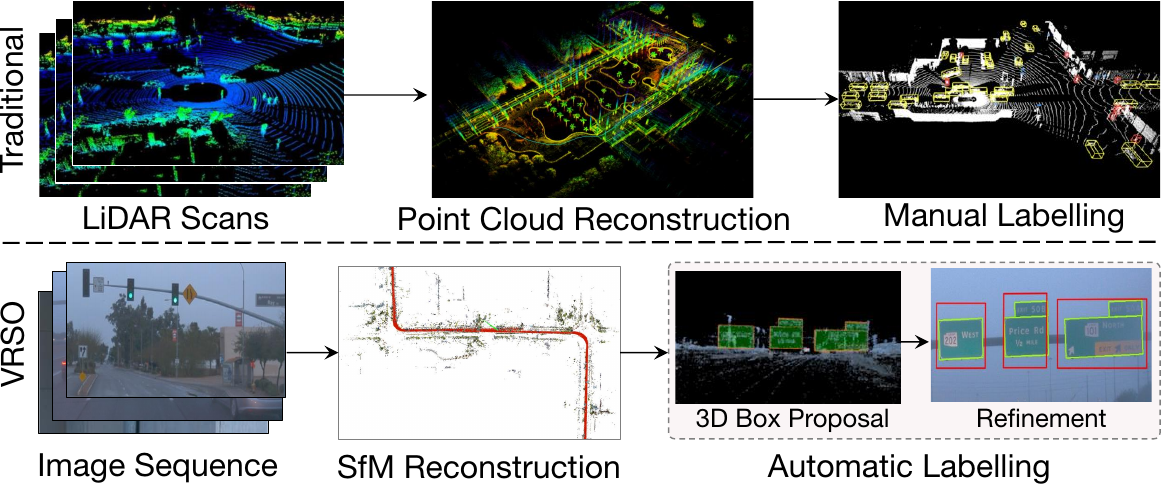}
  \caption{Traditional~\cite{mei2022waymo} and our VRSO annotation pipelines. Our proposed VRSO has lower cost and higher efficiency.}
  \label{fig:compare}
  \vspace{-1.5em}
\end{figure}
Static object detection (SOD), including traffic lights, guideboards, and traffic cones, is critical to modern intelligent driving systems. Most detection algorithms are data-driven, utilizing deep neural networks~\cite{hu2023planning} that demand a high volume of training data. Although existing public driving datasets (e.g., WOD\cite{mei2022waymo} and ZOD\cite{alibeigi2023zenseact}) provide accurate 3D annotations for SOD tasks, their quantities are insufficient for developing a reliable intelligent driving system in complex environments. In practice, high-quality (also quantity) training samples with SOD labels are annotated daily to fix long-tail cases~\cite{yang2021mlife}. Typically, these 3D annotations are manually labelled on point cloud data derived from LiDAR scans, a time-consuming process and prone to errors (Fig.~\ref{fig:compare} (top)). Furthermore, manual labelling struggles to capture the variability and complexity of real-world scenarios, often failing to account for occlusions, varying lighting conditions, and diverse angles of view (yellow arrows in Fig.~\ref{fig:teaser}). These limitations underscore the need for a fully automated SOD reconstruction algorithm capable of handling the intricate demands of real-world intelligent driving applications.

\begin{figure*}[t!]
    \includegraphics[width=1\textwidth]{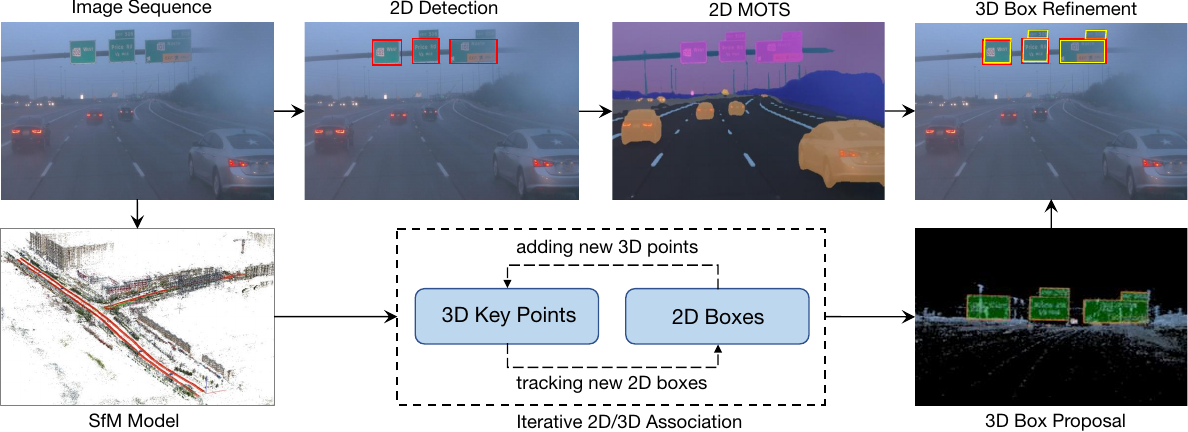}
  \caption{Our proposed VRSO mainly contains six steps: (1) With image sequence, Structure-from-Motion (SfM) obtains accurate ego poses and sparse 3D key points. The 2D-3D correspondence of these key points provides association information across frames. (2) Multi-Object Tracking and Segmentation (MOTS) obtains 2D static objects (detection and segmentation). (3) The initial 3D annotations are proposed based on an Iterative 2D/3D Association of the above 3D points and 2D boxes. (4) These proposals are further refined by splitting or merging in the instance levels. (5) Aiming at the consistency of reprojection, reduce the reprojection error and further optimize the 3D parameters (position, orientation, size). (6) Automatically annotate static objects based on the optimized 3D parameters and tracking information (yellow is VRSO, red is Waymo GT). Best viewed in color.}
  \label{fig:pipeline}
  \vspace{-0.5em}
\end{figure*}
In response to these challenges, we propose VRSO, a \textbf{V}ision-centric \textbf{R}econstruction for \textbf{S}tatic \textbf{O}bject annotation (Fig.~\ref{fig:compare} (bottom)). VRSO leverages information from Structure-from-Motion (SfM), 2D object detection, and instance segmentation results. It begins with 2D object detection for initial proposals. By integrating keypoint-matching features provided by SfM, it can associate 2D object detection boxes across different image frames. Subsequently, the 3D semantic key points can be extracted to generate 3D bounding box proposals. Instance segmentation then offers precise semantic key point locations for final 3D bounding box refinement. Thus, VRSO constructs 2D relationships between instances to improve recognition efficiency. It also uses the Euclidean distance between the 3D vertices and the corresponding 2D projection Intersection over Union (IoU) on the observation image to decide on merging. This approach avoids the re-identification of the same instance, enhancing association efficiency. For static objects, VRSO extracts key points through instance segmentation and contour, addressing the challenges of integration and deduplication of static objects from different viewpoints, and the difficulty of insufficient observation due to occlusion issues, thereby improving annotation accuracy. As illustrated in Fig.~\ref{fig:teaser} (bottom), VRSO demonstrates higher robustness and geometric accuracy than manual labelling results from the Waymo Open Dataset.

Our main contributions are: 1) We propose VRSO, a fully automatic static object reconstruction framework, capable of providing high-quality annotations. 2) Our method's annotations show higher consistency and accuracy in terms of reprojection errors thanks to the vision-based reconstruction pipeline. 3) We conduct extensive experiments on the WOD to demonstrate that our proposed VRSO can provide ground truth labels with accuracy comparable to manual labeling results.
\section{RELATED WORKS}
\label{sec:related}
VRSO can be seen as an automatic SOD labelling system. It involves image-based 3D reconstruction, 2D object detection, instance segmentation (see Fig.~\ref{fig:pipeline}), etc.

\subsection{Structure from Motion for Driving Data}
While Lidar is expensive and unsuitable for all cars, the high-precision attitude of the vehicle is the key to recovering static objects. Since GPS mainly relies on receiving satellite signals to calculate position and speed, the accuracy can only reach a few meters in open space. IMU estimates attitude by measuring acceleration and angular velocity. Its accuracy is affected by integral cumulative errors and will be affected by drift and other problems when used for a long time. However, SfM can provide higher attitude accuracy in a good environment. At the same time, for static scenes and image sequences with a large number of overlapping areas, the accuracy level is higher. The attitude accuracy provided by GPS and IMU is far less than that of SfM, so we choose SfM.

SfM has been developed for decades in the field of computer vision. There are many excellent open-source projects like COLMAP~\cite{schonberger2016structure}, OpenSfM~\cite{adorjan2016opensfm}, HLOC~\cite{sarlin2019coarse}, etc. Some works~\cite{mei2023rome, zhang2023vision} are extensions on specific domains of efficiency and robustness, such as driving scene reconstruction based on COLMAP modification in terms of initialization, feature extraction and bundle adjustment.

\begin{figure*}[t!]
    \includegraphics[width=1\textwidth]{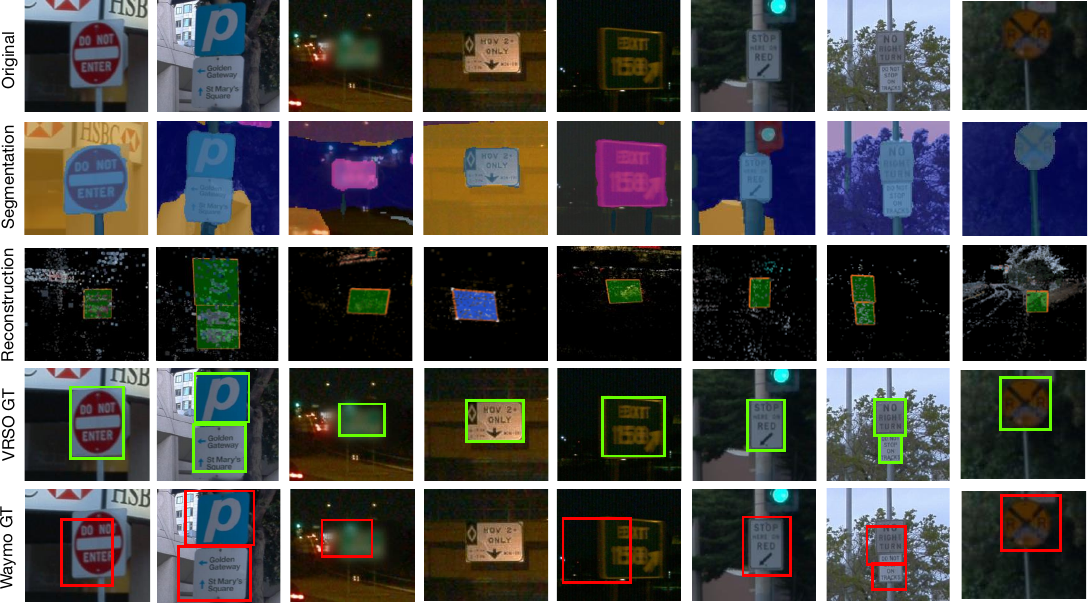}
  \caption{Visualization of annotations. Green and red boxes are the annotations from VRSO and WOD, respectively. We can find that annotations from our proposed VRSO are more accurate to the target than WOD, even in night time with poor illumination conditions.}
  \label{fig:comparison}
  \vspace{-0.5em}
\end{figure*}

\subsection{Object Detection and Segmentation}
The SOD tasks in intelligent driving have gradually transformed from 2D object detection into 3D object detection~\cite{he2020structure}. The algorithms are mainly dominated by deep neural networks~\cite{szegedy2013deep}. These algorithms can be divided into three categories: dual-stage object detection, single-stage object detection, and transformer-based object detection. RCNN~\cite{girshick2015region}, FPN~\cite{lin2017feature}, Faster RCNN~\cite{girshick2015fast} \textit{et al.} are all typical two-stage object detection. YOLO~\cite{redmon2016you},  SSD~\cite{liu2016ssd} and Retinanet~\cite{lin2017focal} \textit{et al.} are typical single-stage object detection. Compared with two-stage object detection~\cite{zhang2020cascaded}, single-stage object detection represented by YOLO~\cite{wang2023improved} achieves high calculation speed while ensuring high prediction accuracy. RelationNet~\cite{hu2018relation} and DETR~\cite{carion2020end} are transformer-based object detection. DETR uses CNN to extract features of images, combined with position coding commonly used in the NLP field, to generate detection data. However, DETR does not use multi-scale features for detection, which leads to long training times and low accuracy of small object detection. While most automatic 3D reconstruction of SOD requires LiDAR point clouds as inputs~\cite{wang2022traffic}, existing methods rarely cover this practical requirement of using only images for 3D SOD annotation.

\subsection{Driving Datasets with SOD 3D Annotation}
The nuScenes dataset~\cite{caesar2020nuscenes} uses a 32-line LiDAR with an acquisition rate of 20Hz and provides detailed temporal annotation for 3D objects at 2Hz. The Waymo Open Dataset~\cite{sun2020scalability} excels in geographic diversity and is three times the size of nuScenes. The ZOD dataset~\cite{alibeigi2023zenseact} is an open dataset for 3D scene understanding. It includes 2D and 3D annotations of different types of traffic signs. However, using LiDAR as input, all SOD ground truth in these datasets are manually (or semi-automatically) created by experienced human annotators using commercial labelling tools, which is time-consuming and laborious. For instance, the WOD requires 1541 standard man-days (8 hours per day) of human labelling for all 12 million 3D bounding boxes (detailed in Section~\ref{sec:exp:efficiency}). It is impractical to apply SOD annotation on daily returned data using the above labelling methods~\cite{yang2021mlife}. Besides, additional effort (time and money) is required to ensure the consistency and accuracy (aka. quality control) of manual annotation~\cite{grosman2020eras}.
\section{Method}
\label{sec:method}
Fig.~\ref{fig:pipeline} illustrates the pipeline of our proposed VRSO algorithm. It consists of two parts: scene reconstruction and static object annotation. Scene reconstruction adopts an SfM-based algorithm to recover the accurate image poses and sparse 3D key points. The static object annotation algorithm is the core of VRSO. First, an off-the-shelf 2D object detection and segmentation algorithm is applied to generate proposals. Second, the 3D-2D key point correspondence from the SfM model is utilized to track the 2D instances across frames. Finally, the reprojection consistency is introduced to refine the parameter of the 3D annotations of static objects. Since VRSO guarantees all 3D elements and their correspondence to 2D objects, 3D-2D correspondence and reprojection accuracy are also guaranteed (or improved) without LiDAR. The pseudo-code of the proposed VRSO pipeline is illustrated in Algorithm 1.

\subsection{Scene Reconstruction}
Following CAMA~\cite{mei2023rome,zhang2023vision}, we adopt COLMAP~\cite{schonberger2016structure} and tailor it to fit the driving scenario. All the surrounding cameras are used to reconstruct the SfM model. To further boost the performance on static object reconstruction, the SuperPoint feature point extractor~\cite{detone2018superpoint} is trained on the sampled datasets that focus on the guideboards, traffic lights, and traffic cones. By optimizing SfM, we achieve roughly five times efficiency boost and 20\% robustness (success rate) improvements for self-driving datasets. The sampling strategy is based on the 2D segmentation results of the original images. The scene reconstruction part provides accurate ego vehicle poses and sparse 3D feature key points for the further static object annotation algorithm.

\begin{figure}[t!]
\vspace{1em}
    \includegraphics[width=1\linewidth]{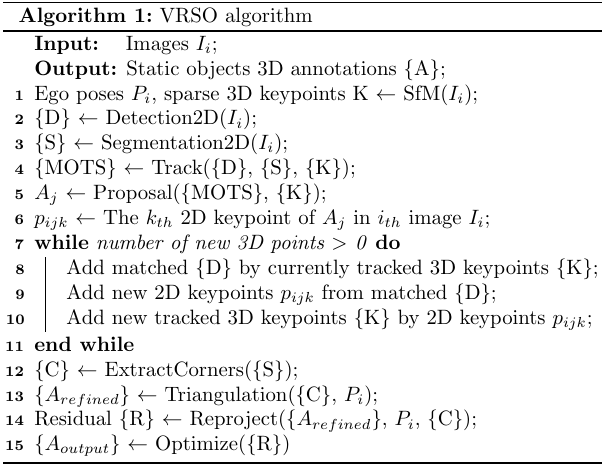}
  \vspace{-2em}
\end{figure}

\subsection{Static Object Annotation}
\subsubsection{Association}
The first step is to obtain 2D detections with instance masks and unique object IDs across frames: a typical Multi-Object Tracking and Segmentation (MOTS)~\cite{voigtlaender2019mots} task. However, existing MOTS algorithms mainly focus on dynamic objects such as vehicles and pedestrians. We then extend MOTS to static objects from segmentation, detection, and association. First, we adopt a 2D detection algorithm~\cite{zong2023detrs} to provide static object classification results, including traffic lights, traffic cones, and guideboards. Since Segment Anything Model  (SAM)~\cite{kirillov2023segment} uses convolutional neural networks for image segmentation, it has powerful feature extraction capabilities. At the same time, SAM supports flexible prompts, real-time calculation of masks and fuzzy cognitive capabilities, which can achieve significant performance improvements in various image segmentation tasks. Therefore we use SAM as an off-the-shelf 2D segmentation algorithm. By combining SAM and detected 2D bounding boxes, the 2D instance segmentation results are obtained. The key points from the SfM are used to build associations of 2D instances across frames. The 2D feature points within each instance mask have association information across frames based on their corresponding 3D key points.

We propose an offline tracking algorithm based on the 3D-2D key point correspondence. In detail, we track static objects in the following four steps: 
\begin{itemize}
    \item \textbf{Step 1:} Extract the 3D points inside the 3D bounding box based on the key points from the SfM model.
    \item \textbf{Step 2:} Calculate the coordinates of each 3D point on the 2D map according to the 2D-3D matching relationship obtained in SfM reconstruction.
    \item \textbf{Step 3:} Determine the corresponding instance of the 3D point on the current 2D map based on the 2D map coordinates and instance segmentation corner points.
    \item \textbf{Step 4:} Determine the correspondence between the 2D observation value of each 2D image and the 3D bounding box formed by its corresponding 3D point coordinates.
\end{itemize}

\subsubsection{Proposal Generation}
In this step, the parameters (location, orientation, size) of 3D bounding boxes of the static objects are initialized for a whole video clip. They can be modified and refined through the rest of the procedures. To recover 3D parameters of static objects from 2D images, the 3D key points from the SfM model can be used as stepping stones. Each key point has an accurate 3D position and correspondence to 2D images. For each 2D instance, the feature points within the 2D instance mask are extracted. Then, a cluster of corresponding 3D key points can be seen as a proposal for the 3D bounding box. We initialized these elements with different vector representations based on the classification information. The guideboard is represented as a rectangle with an orientation in space, and it has 6 degrees of freedom in the Waymo world coordinate system, including translation $(x,y,z)$, orientation ($\theta$), and size (width and height). The traffic light has 7 degrees of freedom, considering its depth. The traffic cone is represented similarly to the traffic light.

\begin{figure}[t!]
    \includegraphics[width=0.48\textwidth]{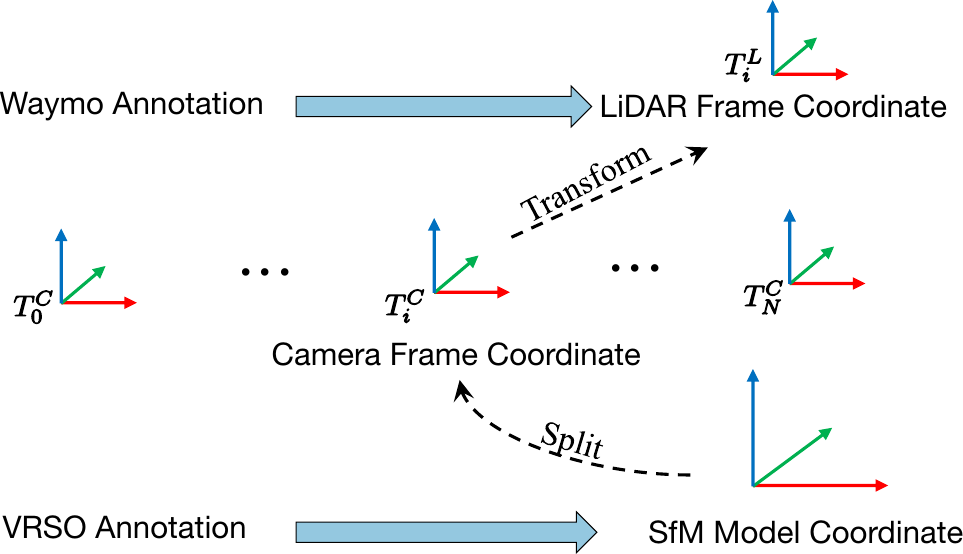}
  \caption{Converting clip-wise VRSO annotation into frame-wise. $T^{C}_{i}$ and $T^{L}_{i}$ denote the camera and LiDAR coordinates at time $t$, respectively. $t=0,...,N$.}
  \label{fig:split}
\end{figure}

\begin{figure*}[t!]
    \includegraphics[width=1\textwidth]{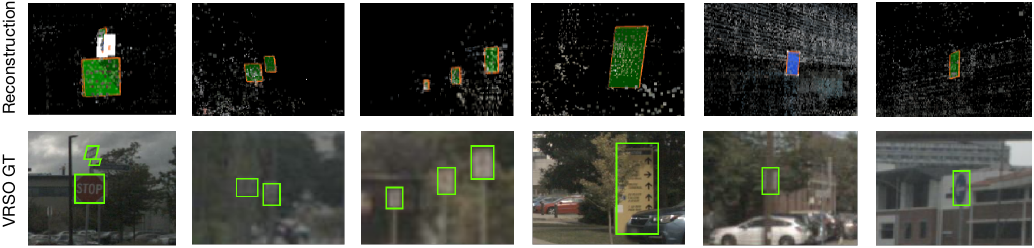}
  \caption{Visualization results of VRSO on the nuScenes dataset. The top row shows the reconstructed guideboards in 3D along with sparse SfM point clouds. The bottom row shows the reprojection results on the images. Our proposed method can be easily applied to different driving dataset.}
  \label{fig:nuscenes}
  \vspace{-0.0em}
\end{figure*}

\subsubsection{Proposal Refinement}
The initial proposals contain mismatches and errors due to insufficient observation. Thus, a refinement procedure is introduced to improve the overall accuracy. It consists of the following steps: 
\begin{itemize}
\item \textbf{Step 1:} Extract the outline of each static object from its 2D instance segmentation. 
\item \textbf{Step 2:} Fit the minimum oriented bounding box (OBB) for the outline contour.
\item \textbf{Step 3:} Extract the vertices of the minimum bounding box.
\item \textbf{Step 4:} Calculate the orientation based on the vertices and centre point and determine the vertex order.
\item \textbf{Step 5:} Generate the minimum bounding box from 3D key point clustering can not guarantee accuracy. There are many cases when multiple signboards are stitched together, or there are some vast signboards with few 3D feature key points, leading to incorrect clustering of 3D key points. So the split and merge procedures are conducted based on the 2D detection and instance segmentation results.
\item \textbf{Step 6:} Detect and reject observations that contain occlusion. The vertex extraction from 2D instance segmentation masks requires that all four corners of each signboard are visible. However, due to occlusions, this requirement cannot always be satisfied. An additional procedure must be introduced to detect these situations and remove these frames from observations. We extract the axis-aligned bounding box (AABB) from the instance segmentation and calculate the area ratio between AABB and 2D detection boxes. If there is no occlusion, these two ways of area calculation should be close. Based on this observation, a threshold $A_th$ is set to detect occlusions. We set the following threshold to detect the occlusion:
\end{itemize}

\begin{equation}
    \frac{Area_{AABB}}{Area_{Det}} < A_{th}
\end{equation}

% \begin{figure}[t!]
% \vspace{1em}
%     \includegraphics[width=1\linewidth]{image/nuscenes6.pdf}
%   \caption{Visualization results of VRSO on the nuScenes dataset.}
%   \label{fig:nuscenes}
%     \vspace{-2em}
% \end{figure}

\subsubsection{Triangulation}
The triangulation is used to obtain the initial vertex value of a static object under 3D conditions. By examining the number of key points in the 3D bounding box obtained by SfM and instance segmentation during scene reconstruction, only the instances whose number of key points is more significant than a threshold are considered stable and valid observations. For these instances, the corresponding 2D bounding boxes are considered valid observations. Through 2D observations of multiple images, the 2D bounding box vertices are triangulated to obtain the coordinates of bounding boxes. Then, the parameters (location, orientation, size) can be easily calculated from the 3D coordinates of these vertices. For circular signboards without distinguished ``bottom left, top left, top right, top right, and bottom right" vertices on the mask, the mask vertex extraction method is not applicable. Thus, we need to identify these circular sign boards. We use 2D detection results as the observation results of circular objects and 2D instance segmentation masks for contour extraction. The centre point and radius are calculated by the least-square fitting algorithm. The parameters of a circular signboard include centre point $(x,y,z)$, orientation ($\theta$) and radius ($r$).

\subsubsection{Tracking Refinements}
The tracking inside the VRSO is based on feature point matching from the SfM model. If the feature points of an instance are insufficient, the tracking results may be unreliable. Thus, the same instance will have multiple tracking IDs, which will decrease the accuracy of annotation. Therefore, we determine whether to merge these separated instances based on the Euclidean distance of the 3D bounding box vertices and the 2D bounding box projection IoU. Once the merging is completed, the 3D feature points inside the instance can be gathered to associate more 2D feature points. As shown in Fig~\ref{fig:pipeline} (iterative 2D/3D association), this loop is considered terminated until none of the 2D feature points can be added.

\subsubsection{Final Parameter Optimization}
The final step is to adjust the annotation parameters by minimizing the residual reprojection errors. Taking a rectangle signboard for example, the optimizable parameters include the location $(x,y,z)$, orientation ($\theta$), and size $(w,h)$, in total six degrees of freedom. The main steps are: 
\begin{itemize}
    \item \textbf{Step 1:} Convert six degrees of freedom into four 3D points and calculate the rotated matrix.
    \item \textbf{Step 2:} Project the converted four 3D points onto the 2D image.
    \item \textbf{Step 3:} Calculate the residual between the projection result and the corner point result from instance segmentation.
    \item \textbf{Step 4:} Update the bounding box parameters in Ceres solvers\cite{Agarwal_Ceres_Solver_2022} with Huber loss~\cite{huber1992robust} to reduce the impact of outliers on the optimization results.
\end{itemize}
After optimization, the residual value can be minimized for more accurate labelling results. Built on that, the projection of the rectangular frame in the image can be closer to the actual observed position of the rectangular frame, thereby improving the accuracy of the bounding boxes.

\section{Experiments}
\label{sec:exp}
We first introduce the dataset used for annotation and evaluation. We also assess the proposed VRSO via a set of quantitative and qualitative comparisons. It should be noted that VRSO has been deployed by a number of intelligent driving companies~\cite{lin2023sparse4d,hu2023planning,wang2023interpretable,pan2023baeformer} in practice. Thus, we do not shy away from discussing the limitations and improvements behind the VRSO system in the last part.

\subsection{Dataset}
\label{sec:exp:dataset}
We use the Waymo Open Dataset (WOD)~\cite{mei2022waymo} and the nuScenes dataset~\cite{caesar2020nuscenes} to verify the effectiveness and accuracy of our proposed VRSO algorithm. The visualization results of VRSO on the nuScenes dataset are shown in Fig.~\ref{fig:nuscenes}. The WOD and the nuScenes dataset provide 3D annotation for static objects, including guideboards, traffic signs, etc. The WOD and the nuScenes dataset are famous for their accurate calibration and time synchronization, which are essential prerequisites to verify reprojection consistency and accuracy.

\noindent \textbf{Properties}: The WOD consists of high-resolution sensor data collected by Waymo’s self-driving vehicles under various conditions. Hardware includes five LiDAR and five cameras. The dataset contains 1150  driving records, totalling 6.4 hours, with an average length of approximately 20 seconds. The entire dataset contains 230 thousand annotated frames, with approximately 12 million 3D bounding boxes and 9.9 million 2D bounding boxes. The nuScenes dataset contains 1000 scenes, each containing 15 seconds of driving data. There are 28130, 6019, and 6008 samples for training, validation, and testing.

\noindent \textbf{Conversion}: The static object annotations provided by the WOD are frame-wise. The annotations are produced by manual labels in LiDAR scans for each frame. While VRSO generates annotations clip-wise, we split VRSO annotations into each frame for a fair comparison and then transform them according to the LiDAR coordinate system with the provided calibration (see Fig.~\ref{fig:split}).

\subsection{Quantitative Evaluation}
\begin{table}[t!]
\normalsize
\centering
    \caption{The computation time (on average) of each VRSO component. The core components of VRSO (proposal and refinement) only cost 22 seconds of computation per clip.}
    \begin{tabular}{c c c c c}
    \hline
    SfM Reconstruction       & MOTS      & Proposal     & Refinement  \\ \hline
    64 mins                  & 4.7 mins  & 20 secs      & 2 secs      \\ \hline
    \end{tabular}
    \label{tab:runtime}
\end{table}

We first report the time cost of each component in VRSO. After that, we evaluate the accuracy and consistency of VRSO by comparisons with WOD in 2D and 3D spaces.

\noindent \textbf{Component Time Cost}: We use more than 67 random clips (around 990 frames per clip) from the WOD dataset for VRSO annotation. The average time cost in each component is detailed in Table~\ref{tab:runtime}. We can clearly find that SfM Reconstruction is the most time-consuming part (64 minutes), followed by MOTS (4.7 minutes). The rest parts cost less than 30 seconds. As we discussed in Section~\ref{sec:exp:dirty}, the SfM reconstruction could be accelerated by parallel acceleration~\cite{zhu2018very} in practice.

\begin{table}[t!]
\normalsize
\centering
\setlength{\tabcolsep}{3.5pt}
    \caption{The quantitative evaluation of VRSO in 2D and 3D spaces. Det denotes the 2D detection results generated by UniDetector~\cite{wang2023detecting} and manual correction. Thus, it can be treated as the ground truth. $E_{2D}$ and $E_{3D}$ denotes the projection error in 2D (pixels) and 3D (meters) spaces.}
\renewcommand{\arraystretch}{1.1}
\begin{tabular}{c|cccccc}
\hline
 & Prediction & Reference & Precision & Recall & $E_{3D}$ & $E_{2D}$ \\ \hline
\multirow{2}{*}{2D} & Waymo & Det & 0.67 & 0.35 & - & 9.98 \\
 & VRSO & Det & \textbf{0.88} & \textbf{0.82} & - & \textbf{2.13} \\ \hline
3D & VRSO & Waymo & 0.60 & 0.24 & 0.27 & - \\ \hline
\end{tabular}
    \label{tab:compare}
\end{table}
\noindent \textbf{Quality in 2D Space}: We compare the reprojection consistency between our proposed VRSO and WOD. To do so, five steps are applied for the reprojection:
\begin{itemize}
    \item \textbf{Step 1:} Reproject the 3D annotation vector elements into each 2D image.
    \item \textbf{Step 2:} Use off-the-shelf instance segmentation techniques to extract all instances in the image space and fit a polyline to each instance.
    \item \textbf{Step 3:} Use the Hungarian algorithm~\cite{kuhn1955hungarian} to match the projected elements from Step 1 with the extracted elements from Step 2.
    \item \textbf{Step 4:} Calculate the precision and recall following the 2D detection metrics. The matching IoU threshold is set to 0.5.
\end{itemize}
The evaluation results are reported in Table~\ref{tab:compare} (top). We find that our proposed VRSO yields higher precision and recall. Moreover, the projection error is only 2.13 pixels on average, which is around 4.6 times lower than the Waymo annotation.

\noindent \textbf{Quality in 3D Space}:We directly compare our annotations against WOD in 3D space. Note that annotations of WOD in 3D space are considered ground truth for the 3D object detection tasks. Without losing generality, we take WOD annotation as a reference and VRSO annotation as a prediction. The association of 3D bounding boxes is measured by the Euclidean centre point distance. The threshold is empirically set to 1 meter. Similarly, we calculate the precision, recall and position error $E_{3D}$. As detailed in Table~\ref{tab:compare} (bottom), our VRSO annotation is around 0.27 meters from the manual labelled Waymo. In fact, as can be seen from Fig.~\ref{fig:comparison}, our VRSO annotation is more perception-friendly, tight, and accurate to the ground truth.

\begin{figure}[t!]
    \includegraphics[width = 0.49\textwidth]{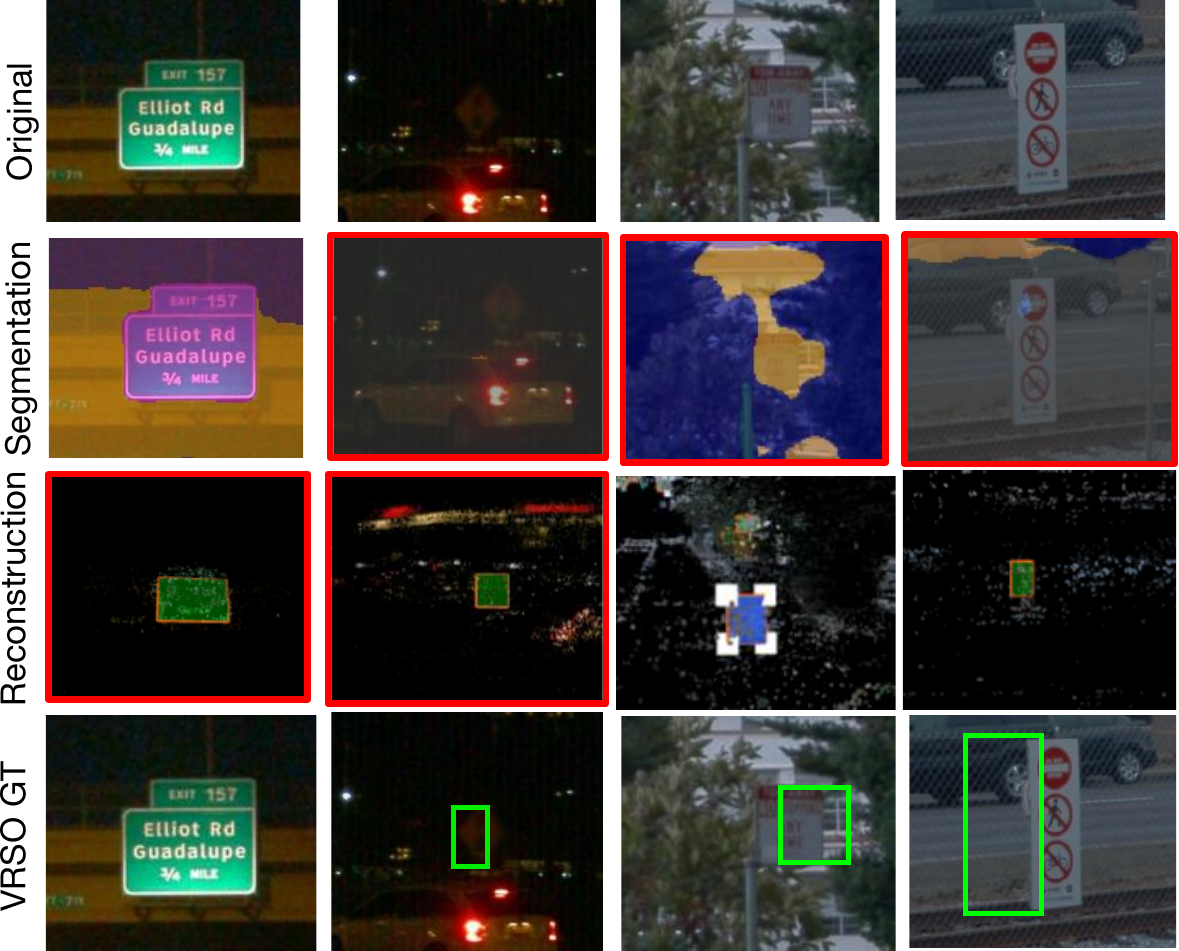}
    \caption{Long-tail cases from harsh environments of Waymo Open Dataset. The intermediate results are drawn to show the failure reasons.}
    \label{fig:long-tail}
\end{figure}

\subsection{Qualitative Evaluation}
\label{sec:exp:dataset:qualitative}
To visually compare the consistency and accuracy between VRSO and WOD, we reproject the 3D bounding boxes back to the images in Fig.~\ref{fig:comparison} (two bottom rows). Annotations from VRSO and WOD are marked in green and red. In addition, the intermediate segmentation and reconstruction results from our proposed VRSO are also presented in the corresponding columns. For WOD annotation (red boxes), we observe that (1) the boxes do not correctly cover the target in the first and fifth columns, (2) the guideboard in the fourth column is missed, and (3) the rest of the boxes do not tightly cover the target. However, annotations from our proposed VRSO (green boxes) are more consistent and accurate (good adhesion and minimal omission of labels). Furthermore, the second and seventh columns indicate that our method can properly annotate multiple adjacent sign boards.

\subsection{Efficiency Comparison}
\label{sec:exp:efficiency}
The Waymo Open Dataset contains 12 million 3D bounding boxes and 9.9 million 2D bounding boxes. With the help of pre-trained detection models, the average manual labelling speed is 3.7 seconds for 3D bounding boxes and 1.5 seconds for 2D bounding boxes\cite{lee2018leveraging}. Given eight working hours daily, it takes about 1541 man-days of human labelling for all 12 million 3D bounding boxes. In comparison, our method takes about 4.9 minutes of computation for each driving clip. It costs 95 hours of calculation for all 1150 clips, which is more than 16x speed up than human annotations.

\subsection{Limitations and Improvements}
\label{sec:exp:dirty}

Fig.~\ref{fig:long-tail} illustrates some results in challenging conditions, such as extremely low resolution and insufficient lighting that affect the loss of objects during reconstruction or segmentation. SfM accuracy is indeed affected under low light and exposure conditions, but the impact is limited. To alleviate this problem, we take the following measures to improve the annotation quality: (1) The camera trigger time is used as the timestamp of each frame. (2) For camera-LiDAR frame pairs with timestamp differences greater than 10ms, they are removed from the evaluation for fair comparison. (3) The parameter $A_{th}$ can be empirically set to around 0.95 to ensure the accuracy of annotating occluded objects. (4) When performing 3D reconstruction, SfM uses local reconstruction and global merging methods to reconstruct multiple images in parallel. Based on these measures, the annotation quality of VRSO is improved under low light and exposure conditions.

\section{Conclusion}
\label{sec:conclusion}
This paper introduces VRSO, a novel framework that achieves high-precision and consistent 3D annotations for static objects. VRSO cohesively integrates detection, segmentation, and Structure from Motion (SfM) algorithms, significantly easing integration challenges within modern intelligent driving auto-labelling systems. Notably, VRSO eliminates the necessity for human intervention in the annotation process, delivering results that are on par with manual annotations in LiDAR scans. We have conducted qualitative and quantitative evaluations using the widely recognised Waymo Open Dataset to demonstrate its effectiveness and practical applicability: around 16x speed up than human annotation, while maintaining optimal consistency and accuracy.

\bibliographystyle{IEEEtran}
\bibliography{ref}

\begin{thebibliography}{10}
\providecommand{\url}[1]{#1}
\csname url@samestyle\endcsname
\providecommand{\newblock}{\relax}
\providecommand{\bibinfo}[2]{#2}
\providecommand{\BIBentrySTDinterwordspacing}{\spaceskip=0pt\relax}
\providecommand{\BIBentryALTinterwordstretchfactor}{4}
\providecommand{\BIBentryALTinterwordspacing}{\spaceskip=\fontdimen2\font plus
\BIBentryALTinterwordstretchfactor\fontdimen3\font minus \fontdimen4\font\relax}
\providecommand{\BIBforeignlanguage}[2]{{%
\expandafter\ifx\csname l@#1\endcsname\relax
\typeout{** WARNING: IEEEtran.bst: No hyphenation pattern has been}%
\typeout{** loaded for the language `#1'. Using the pattern for}%
\typeout{** the default language instead.}%
\else
\language=\csname l@#1\endcsname
\fi
#2}}
\providecommand{\BIBdecl}{\relax}
\BIBdecl

\bibitem{mei2022waymo}
J.~Mei, A.~Z. Zhu, X.~Yan, H.~Yan, S.~Qiao, L.-C. Chen, and H.~Kretzschmar, ``Waymo open dataset: Panoramic video panoptic segmentation,'' in \emph{European Conference on Computer Vision}.\hskip 1em plus 0.5em minus 0.4em\relax Springer, 2022, pp. 53--72.

\bibitem{hu2023planning}
Y.~Hu, J.~Yang \emph{et~al.}, ``Planning-oriented autonomous driving,'' in \emph{IEEE Conference on Computer Vision and Pattern Recognition}, 2023, pp. 17\,853--17\,862.

\bibitem{alibeigi2023zenseact}
M.~Alibeigi, W.~Ljungbergh, A.~Tonderski, G.~Hess, A.~Lilja, C.~Lindstrom, D.~Motorniuk, J.~Fu, J.~Widahl, and C.~Petersson, ``Zenseact open dataset: A large-scale and diverse multimodal dataset for autonomous driving,'' in \emph{IEEE International Conference on Computer Vision}, 2023.

\bibitem{yang2021mlife}
C.~Yang, W.~Wang, Y.~Zhang, Z.~Zhang, L.~Shen, Y.~Li, and J.~See, ``Mlife: A lite framework for machine learning lifecycle initialization,'' \emph{Machine Learning}, vol. 110, pp. 2993--3013, 2021.

\bibitem{schonberger2016structure}
J.~L. Schonberger and J.-M. Frahm, ``Structure-from-motion revisited,'' in \emph{IEEE Conference on Computer Vision and Pattern Recognition}, 2016, pp. 4104--4113.

\bibitem{adorjan2016opensfm}
M.~Adorjan, ``Opensfm: A collaborative structure-from-motion system,'' Ph.D. dissertation, Wien, 2016.

\bibitem{sarlin2019coarse}
P.-E. Sarlin, C.~Cadena, R.~Siegwart, and M.~Dymczyk, ``From coarse to fine: Robust hierarchical localization at large scale,'' in \emph{IEEE Conference on Computer Vision and Pattern Recognition}, 2019, pp. 12\,716--12\,725.

\bibitem{mei2023rome}
R.~Mei, W.~Sui, J.~Zhang, Q.~Zhang, T.~Peng, and C.~Yang, ``Rome: Towards large scale road surface reconstruction via mesh representation,'' \emph{arXiv preprint arXiv:2306.11368}, 2023.

\bibitem{zhang2023vision}
J.~Zhang, S.~Chen, H.~Yin, R.~Mei, X.~Liu, C.~Yang, Q.~Zhang, and W.~Sui, ``A vision-centric approach for static map element annotation,'' \emph{arXiv preprint arXiv:2309.11754}, 2023.

\bibitem{he2020structure}
C.~He, H.~Zeng, J.~Huang, X.-S. Hua, and L.~Zhang, ``Structure aware single-stage 3d object detection from point cloud,'' in \emph{IEEE Conference on Computer Vision and Pattern Recognition}, 2020, pp. 11\,873--11\,882.

\bibitem{szegedy2013deep}
C.~Szegedy, A.~Toshev, and D.~Erhan, ``Deep neural networks for object detection,'' \emph{Advances in Neural Information Processing Systems}, vol.~26, 2013.

\bibitem{girshick2015region}
R.~Girshick, J.~Donahue, T.~Darrell, and J.~Malik, ``Region-based convolutional networks for accurate object detection and segmentation,'' \emph{IEEE Transactions on Pattern Analysis and Machine Intelligence}, vol.~38, no.~1, pp. 142--158, 2015.

\bibitem{lin2017feature}
T.-Y. Lin, P.~Doll{\'a}r, R.~Girshick, K.~He, B.~Hariharan, and S.~Belongie, ``Feature pyramid networks for object detection,'' in \emph{IEEE Conference on Computer Vision and Pattern Recognition}, 2017, pp. 2117--2125.

\bibitem{girshick2015fast}
R.~Girshick, ``Fast r-cnn,'' in \emph{IEEE International Conference on Computer Vision}, 2015, pp. 1440--1448.

\bibitem{redmon2016you}
J.~Redmon, S.~Divvala, R.~Girshick, and A.~Farhadi, ``You only look once: Unified, real-time object detection,'' in \emph{IEEE Conference on Computer Vision and Pattern Recognition}, 2016, pp. 779--788.

\bibitem{liu2016ssd}
W.~Liu, D.~Anguelov, D.~Erhan, C.~Szegedy, S.~Reed, C.-Y. Fu, and A.~C. Berg, ``Ssd: Single shot multibox detector,'' in \emph{European Conference on Computer Vision}.\hskip 1em plus 0.5em minus 0.4em\relax Springer, 2016, pp. 21--37.

\bibitem{lin2017focal}
T.-Y. Lin, P.~Goyal, R.~Girshick, K.~He, and P.~Doll{\'a}r, ``Focal loss for dense object detection,'' in \emph{IEEE International Conference on Computer Vision}, 2017, pp. 2980--2988.

\bibitem{zhang2020cascaded}
J.~Zhang, Z.~Xie, J.~Sun, X.~Zou, and J.~Wang, ``A cascaded r-cnn with multiscale attention and imbalanced samples for traffic sign detection,'' \emph{IEEE access}, vol.~8, pp. 29\,742--29\,754, 2020.

\bibitem{wang2023improved}
J.~Wang, Y.~Chen, Z.~Dong, and M.~Gao, ``Improved yolov5 network for real-time multi-scale traffic sign detection,'' \emph{Neural Computing and Applications}, vol.~35, no.~10, pp. 7853--7865, 2023.

\bibitem{hu2018relation}
H.~Hu, J.~Gu, Z.~Zhang, J.~Dai, and Y.~Wei, ``Relation networks for object detection,'' in \emph{IEEE Conference on Computer Vision and Pattern Recognition}, 2018, pp. 3588--3597.

\bibitem{carion2020end}
N.~Carion, F.~Massa, G.~Synnaeve, N.~Usunier, A.~Kirillov, and S.~Zagoruyko, ``End-to-end object detection with transformers,'' in \emph{European Conference on Computer Vision}.\hskip 1em plus 0.5em minus 0.4em\relax Springer, 2020, pp. 213--229.

\bibitem{wang2022traffic}
M.~Wang, R.~Liu, J.~Yang, X.~Lu, J.~Yu, and H.~Ren, ``Traffic sign three-dimensional reconstruction based on point clouds and panoramic images,'' \emph{The Photogrammetric Record}, vol.~37, no. 177, pp. 87--110, 2022.

\bibitem{caesar2020nuscenes}
H.~Caesar, V.~Bankiti, A.~H. Lang, S.~Vora, V.~E. Liong, Q.~Xu, A.~Krishnan, Y.~Pan, G.~Baldan, and O.~Beijbom, ``nuscenes: A multimodal dataset for autonomous driving,'' in \emph{IEEE Conference on Computer Vision and Pattern Recognition}, 2020, pp. 11\,621--11\,631.

\bibitem{sun2020scalability}
P.~Sun, H.~Kretzschmar, X.~Dotiwalla, A.~Chouard, V.~Patnaik, P.~Tsui, J.~Guo, Y.~Zhou, Y.~Chai, B.~Caine \emph{et~al.}, ``Scalability in perception for autonomous driving: Waymo open dataset,'' in \emph{IEEE Conference on Computer Vision and Pattern Recognition}, 2020, pp. 2446--2454.

\bibitem{grosman2020eras}
J.~S. Grosman, P.~H. Furtado, A.~M. Rodrigues, G.~G. Schardong, S.~D. Barbosa, and H.~C. Lopes, ``Eras: Improving the quality control in the annotation process for natural language processing tasks,'' \emph{Information Systems}, vol.~93, p. 101553, 2020.

\bibitem{detone2018superpoint}
D.~DeTone, T.~Malisiewicz, and A.~Rabinovich, ``Superpoint: Self-supervised interest point detection and description,'' in \emph{IEEE Conference on Computer Vision and Pattern Recognition}, 2018, pp. 224--236.

\bibitem{voigtlaender2019mots}
P.~Voigtlaender, M.~Krause, A.~Osep, J.~Luiten, B.~B.~G. Sekar, A.~Geiger, and B.~Leibe, ``Mots: Multi-object tracking and segmentation,'' in \emph{IEEE Conference on Computer Vision and Pattern Recognition}, 2019, pp. 7942--7951.

\bibitem{zong2023detrs}
Z.~Zong, G.~Song, and Y.~Liu, ``Detrs with collaborative hybrid assignments training,'' in \emph{IEEE International Conference on Computer Vision}, 2023, pp. 6748--6758.

\bibitem{kirillov2023segment}
A.~Kirillov, E.~Mintun, N.~Ravi, H.~Mao, C.~Rolland, L.~Gustafson, T.~Xiao, S.~Whitehead, A.~C. Berg, W.-Y. Lo \emph{et~al.}, ``Segment anything,'' in \emph{IEEE International Conference on Computer Vision}, 2023, pp. 4015--4026.

\bibitem{Agarwal_Ceres_Solver_2022}
\BIBentryALTinterwordspacing
S.~Agarwal, K.~Mierle, and T.~C.~S. Team, ``{Ceres Solver},'' 10 2023. [Online]. Available: \url{https://github.com/ceres-solver/ceres-solver}
\BIBentrySTDinterwordspacing

\bibitem{huber1992robust}
P.~J. Huber, ``Robust estimation of a location parameter,'' in \emph{Breakthroughs in Statistics}.\hskip 1em plus 0.5em minus 0.4em\relax Springer, 1992, pp. 492--518.

\bibitem{lin2023sparse4d}
X.~Lin, Z.~Pei, T.~Lin, L.~Huang, and Z.~Su, ``Sparse4d v3: Advancing end-to-end 3d detection and tracking,'' \emph{arXiv preprint arXiv:2311.11722}, 2023.

\bibitem{wang2023interpretable}
B.~Wang, Z.~Wang, C.~Zhu, Z.~Zhang, Z.~Wang, P.~Lin, J.~Liu, and Q.~Zhang, ``Interpretable motion planner for urban driving via hierarchical imitation learning,'' in \emph{International Conference on Intelligent Robots and Systems}, 2023, pp. 1691--1696.

\bibitem{pan2023baeformer}
C.~Pan, Y.~He, J.~Peng, Q.~Zhang, W.~Sui, and Z.~Zhang, ``Baeformer: Bi-directional and early interaction transformers for bird's eye view semantic segmentation,'' in \emph{IEEE Conference on Computer Vision and Pattern Recognition}, 2023, pp. 9590--9599.

\bibitem{zhu2018very}
S.~Zhu, R.~Zhang, L.~Zhou, T.~Shen, T.~Fang, P.~Tan, and L.~Quan, ``Very large-scale global sfm by distributed motion averaging,'' in \emph{IEEE Conference on Computer Vision and Pattern Recognition}, 2018, pp. 4568--4577.

\bibitem{wang2023detecting}
Z.~Wang, Y.~Li, X.~Chen, S.-N. Lim, A.~Torralba, H.~Zhao, and S.~Wang, ``Detecting everything in the open world: Towards universal object detection,'' in \emph{IEEE Conference on Computer Vision and Pattern Recognition}, 2023, pp. 11\,433--11\,443.

\bibitem{kuhn1955hungarian}
H.~W. Kuhn, ``The hungarian method for the assignment problem,'' \emph{Naval research logistics quarterly}, vol.~2, no. 1-2, pp. 83--97, 1955.

\bibitem{lee2018leveraging}
J.~Lee, S.~Walsh, A.~Harakeh, and S.~L. Waslander, ``Leveraging pre-trained 3d object detection models for fast ground truth generation,'' in \emph{International Conference on Intelligent Transportation Systems}, 2018, pp. 2504--2510.

\end{thebibliography}

\end{document}